\documentclass[10pt,twocolumn,letterpaper]{article}

\usepackage{iccv}
\usepackage{times}
\usepackage{epsfig}
\usepackage{graphicx}
\usepackage{amsmath}
\usepackage{amssymb}
\usepackage{booktabs}

\usepackage{comment}
\usepackage{multirow}
\usepackage{cite}
\usepackage{xspace}

\usepackage{caption}
\usepackage{subcaption}
\usepackage{placeins}

\newcommand{\reals}{\mathbb{R}}

\usepackage[pagebackref=true,breaklinks=true,letterpaper=true,colorlinks,bookmarks=false]{hyperref}

\iccvfinalcopy %

\def\iccvPaperID{6336} %

\ificcvfinal\fi

\begin{document}

\title{TextMesh: Generation of Realistic 3D Meshes From Text Prompts}

\author{Christina Tsalicoglou$^{1, 2*}$ \qquad 
Fabian Manhardt$^2$ \qquad Alessio Tonioni$^2$ \\
Michael Niemeyer$^2$ \qquad  Federico Tombari$^{2, 3}$ \\ \\
$^1$ETH Zurich  \qquad $^2$Google \qquad $^3$Technical University of Munich \\
\small{\texttt{ctsalico@ethz.ch} \qquad \texttt{\{fabianmanhardt, alessiot, mniemeyer, tombari\}@google.com}} %
}

\maketitle
\ificcvfinal\thispagestyle{empty}\fi

\begin{abstract}
The ability to generate highly realistic 2D images from mere text prompts has recently made huge progress in terms of speed and quality, thanks to the advent of image diffusion models. %
Naturally, the question arises if this can be also achieved in the generation of 3D content from such text prompts. %
To this end, a new line of methods recently emerged trying to harness diffusion models, trained on 2D images, for supervision of 3D model generation using view dependent prompts. While achieving impressive results, these methods, however, have two major drawbacks. First, rather than commonly used 3D meshes, they instead generate neural radiance fields (NeRFs), making them impractical for most real applications. %
Second, %
these approaches tend to produce over-saturated models, giving the output a cartoonish looking effect. Therefore, in this work we propose a novel method for generation of highly realistic-looking 3D meshes. To this end, we extend NeRF to employ an SDF backbone, %
leading to improved 3D mesh extraction. %
In addition, we propose a novel way %
to finetune the mesh texture, removing the effect of high saturation and improving the details of the output 3D mesh.
\end{abstract}

\vspace{-2mm}
{\let\thefootnote\relax\footnote{{$^*$This work was conducted during an internship at Google.\vspace{-2mm}}}} 
\vspace{-2mm}

\section{Introduction}

Generating photorealistic 2D images from simple text prompts is a rapidly growing field. Thanks to diffusion models and the availability of huge amount of training data with text-image pairs, current models can generate very high-quality images~\cite{saharia2022photorealistic, rombach2022high, ramesh2022hierarchical}. Naturally, the question arises if the same high quality generative capabilities can be achieved for 3D modeling. Unfortunately, this is a much more challenging field as the output space is significantly larger, 3D consistency is required, and there is a lack of large amount of training data pairs for text and 3D models. 

\begin{figure}[t!]
    \centering
    \includegraphics[width=0.47\textwidth]{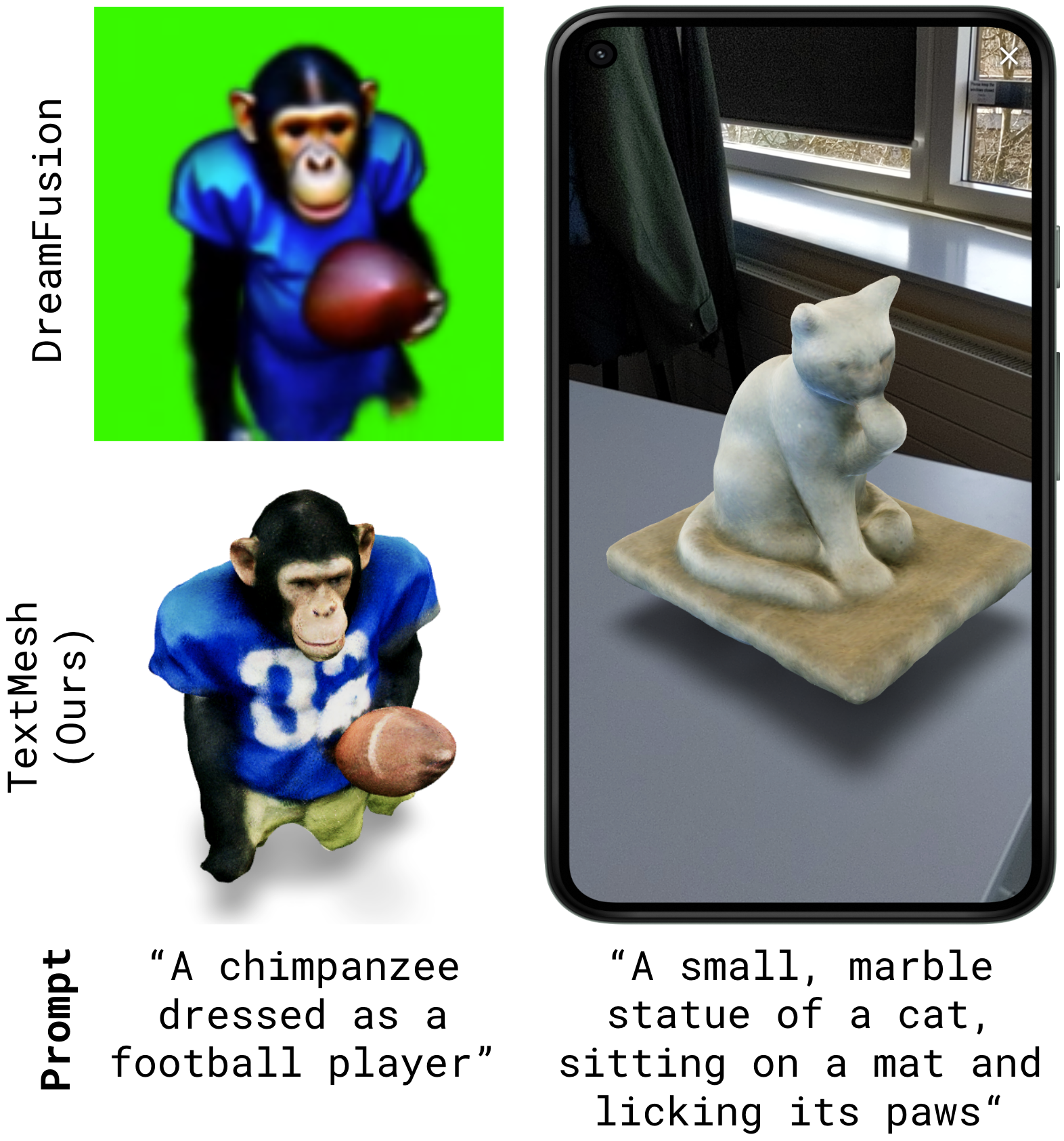}
    \caption{\textbf{Exemplary results} of our TextMesh. Left: We compare our final mesh with the corresponding rendering from the public DreamFusion~\cite{poole2022dreamfusion} gallery. While the results of DreamFusion are overly saturated, almost having a 'cartoonish' appearance, our mesh is more detailed and showcases a more realistic and natural appearance. Right: Since our method estimates a 3D mesh for the prompt instead of a NeRF-like representation, the obtained meshes can be directly plugged into standard computer graphics pipeline to \textit{e.g.} enable AR/VR experiences.}
    \label{fig:teaser}
\end{figure}

\begin{figure*}[t!]
    \centering
    \includegraphics[width=1.\linewidth]{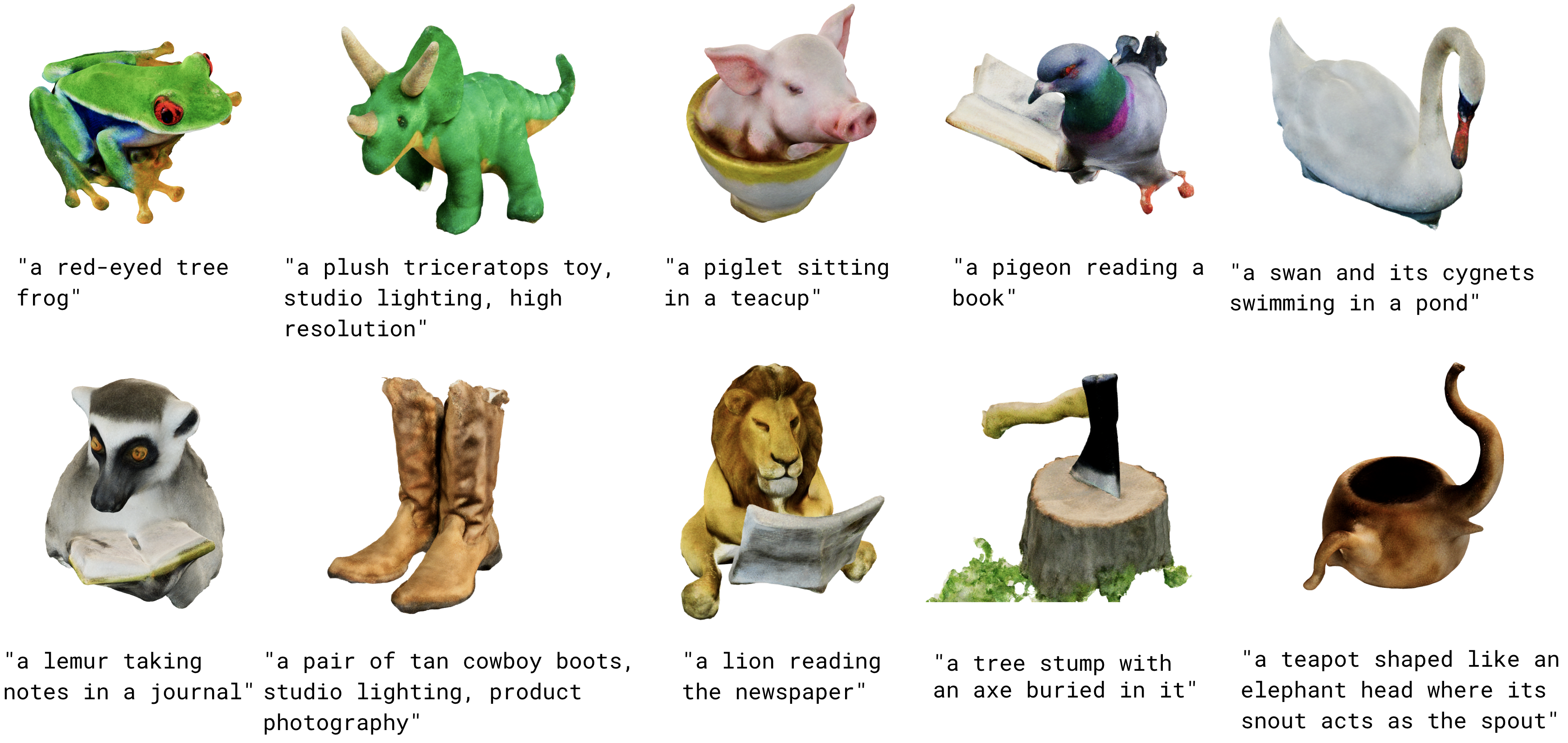}
    \caption{\textbf{Qualitative Results.} Several qualitative 3D meshes generated from the given text prompts. The colors of the meshes are very natural, not showing any over-saturation effects.}
    \label{fig:qual_results}
\end{figure*}

Early methods mostly attempted at deforming template shapes, such as spheres, using a CLIP~\cite{radford2021learning} objective. However, their emerging 3D shapes were still very unsatisfactory in geometry as well as appearance~\cite{mohammad2022clip, michel2022text2mesh, jain2022zero}. To overcome this limitation, DreamFusion~\cite{poole2022dreamfusion} has recently proposed to harness the power of the aforementioned text-to-image diffusion models (\textit{i.e.}~Imagen~\cite{saharia2022photorealistic}) to supervise 3D modelling from text prompts. To this end, they propose to train a Neural Radiance Field (NeRF) with a novel Score Distillation Sampling (SDS) gradient, together with view-dependant prompts. Despite their proposed method being capable of generating impressive results, it still has several downsides. First, the method has a tendency to produce objects with over-saturated colors due to the strong guidance required to make the model converge. Although prompt-engineering, \textit{e.g.} prefixing "\emph{A DSLR photo of [...]}" to the prompt, can mitigate this issue to some extent, the results are still not very satisfactory when it comes to actual realism. Second, \cite{poole2022dreamfusion} represents the 3D scene in the form of a NeRF, which renders the approach impractical to be used within standard computer graphics pipelines. Note that, while it is indeed possible to extract a mesh from NeRF, it is a non-trivial process given the density-based representation~\cite{yariv2021volume,tang2022nerf2mesh}.

In this work, we present TextMesh, a novel method for 3D shape generation from text prompts, targeted at tackling the aforementioned limitations, \textit{i.e.}~generating photorealistic 3D content in the form of standard 3D meshes. As demonstrated in Figure~\ref{fig:teaser}, our generated 3D meshes significantly improve upon \cite{poole2022dreamfusion} for realism and can be directly utilized within standard computer graphics pipelines and applications in AR or VR. 
To accomplish this, we modify DreamFusion to model radiance in the form of a signed distance function (SDF), allowing by design easy extraction of the surface as the 0-level set of the obtained volume. %
Furthermore, in an effort to enhance the mesh quality, we re-texture the output by leveraging another diffusion model, conditioned on color and depth from the mesh. To this end, we render the object from multiple viewpoints and use diffusion to guide texture optimization to enhance realism and details. Nevertheless, when processing individual views independently, the refined texture exhibits severe inconsistencies. Therefore, we propose to run several views simultaneously through the diffusion model instead. To obtain the final texture, we then train on the produced output views together with Score Distillation Sampling to ensure smooth transitions. In Figure~\ref{fig:qual_results}, we illustrate several meshes generated by our proposed method using different prompts.

\noindent To summarize, we propose the following contributions: \textbf{i}) We modify DreamFusion to model radiance in the form of SDF to tailor the model towards mesh extraction. \textbf{ii}) We propose a novel multi-view consistent and mesh conditioned re-texturing, enabling the generation of photorealistic 3D mesh models. \textbf{iii}) We experimentally show that our obtained meshes are geometrically of high quality and showcase more natural textures than the current state-of-the-art, whilst being ready to be deployed into pre-existing graphics pipelines.

\section{Related work}

\paragraph{3D Reconstruction with Neural Fields.}
Traditional 3D reconstruction methods~\cite{Agrawal2001CVPR,Bonet1999ICCV,Kutulakos2000IJCV,Broadhurst2001ICCV,Kutulakos2000IJCV,Seitz1997CVPR} usually rely on underlying depth~\cite{Bleyer2011BMVC,Schoenberger2016ECCV}, or voxel\cite{Agrawal2001CVPR,Bonet1999ICCV,Seitz1997CVPR}-based representations and perform some form of feature matching to fuse multi-view observations to a coherent 3D representation. While leading to satisfactory results in dense multi-view stereo setups, these systems often fail in less constrained scenarios and cannot be integrated easily into other learning-based systems.
In contrast, recent advances in neural fields have achieved impressive results on a variety of tasks.
While seminal works focused on 3D reconstruction from 3D supervision~\cite{Mescheder2019CVPR,Park2019CVPR,Chen2019CVPR}, later works proposed surface rendering techniques~\cite{Niemeyer2020CVPR,Yariv2020NIPS} that require 2D supervision in the form of image and mask data. The introduction of Neural Radiance Fields (NeRFs)~\cite{Mildenhall2020ECCV} enabled impressive view synthesis from only image input via volume rendering. In many downstream applications, however, mesh-based representations are required, and directly extracting a mesh from a NeRF representation is non-trivial\cite{tang2022nerf2mesh}. 
As a result, recent approaches~\cite{yariv2021volume,Oechsle2021ICCV,Wang2021NEURIPS} combine surface and volume rendering techniques to enable mesh extraction from image input.
The goal of this work is to optimize a high-quality mesh and texture from text input. To this end, we adopt the VolSDF~\cite{yariv2021volume} representation due to its state-of-the-art performance and simple design.

\paragraph{Photorealistic Image Generation From Text Prompts.}
Text-to-image models have recently achieved impressive high-fidelity and flexible image synthesis. The huge boost in quality has been made possible by the availability of extremely large datasets of image-text pairs~\cite{schuhmann2022laion} and scalable generator architectures based on diffusion models~\cite{rombach2022high, ramesh2022hierarchical, saharia2022photorealistic, balaji2022ediffi} and transformers~\cite{chang2023muse,yu2022scaling}.
One benefit of diffusion models over other classes of generative models is their flexibility with respect to the conditioning used in the image generation process since all of them support conditioning on text and a seed image. Recent works extend this support to text and a depth map~\cite{rombach2022high} or text and a scene layout~\cite{gafni2022make}. 
In this work, we utilize large-scale pretrained text-to-image models to enable text-to-3D mesh synthesis.

\paragraph{3D Generation From Text Prompts.}
There are a handful of works that attempt to generate 3D objects from text prompts. While most of them use a CLIP objective to supervise generation, a very new direction started to also incorporate large text-to-image diffusion models for training.
As for the former, CLIPMesh~\cite{mohammad2022clip} deforms a 3D sphere using a CLIP loss to obtain a 3D mesh that fits the input prompt, while Text2Mesh~\cite{michel2022text2mesh} similarly uses the CLIP loss to deform a given mesh and adjust its colors to better match the prompts. DreamFields~\cite{jain2022zero} proposes to train a NeRF by rendering it from multiple viewpoints also using a CLIP objective. While these methods can indeed perform 3D object generation from text prompts, their results are very unsatisfactory when it comes to geometry and colors. Hence, more recently, DreamFusion started to investigate how to leverage pre-trained text-to-image diffusion models for 3D generation~\cite{poole2022dreamfusion}. Similar to \cite{jain2022zero}, DreamFusion  trains a NeRF by rendering it from multiple viewpoints, however, supervising the model with their proposed Score Distillation Sampling (SDS) gradient based on \emph{imagen}, a large text-to-image diffusion model~\cite{saharia2022photorealistic}. While leading to impressive results, DreamFusion generates volumetric representations instead of meshes, making it impractical for many downstream applications such as graphics, where standard 3D representations such as meshes are required. Further,  due to the high guidance weights required for optimization, their results tend to be oversaturated rather than photorealistic. To increase the texture resolution and improve the mesh extraction, the concurrent work Magic3D~\cite{lin2022magic3d} proposes a two step approach. Firstly, they use a Dreamfusion like optimization with an SDF representation obtained from the density by subtracting a constant value. Secondly, they extract a mesh and employ differentiable rendering to further optimize it with a SDS objective.

In this work, we propose to overcome the aforementioned limitations by modifying the underlying neural field to represent an SDF instead of a radiance field to make the model better suited for mesh extraction and additionally propose a novel texture refinement to increase photorealism.

\begin{figure*}[t!]
    \centering
    \includegraphics[width=1.\textwidth]{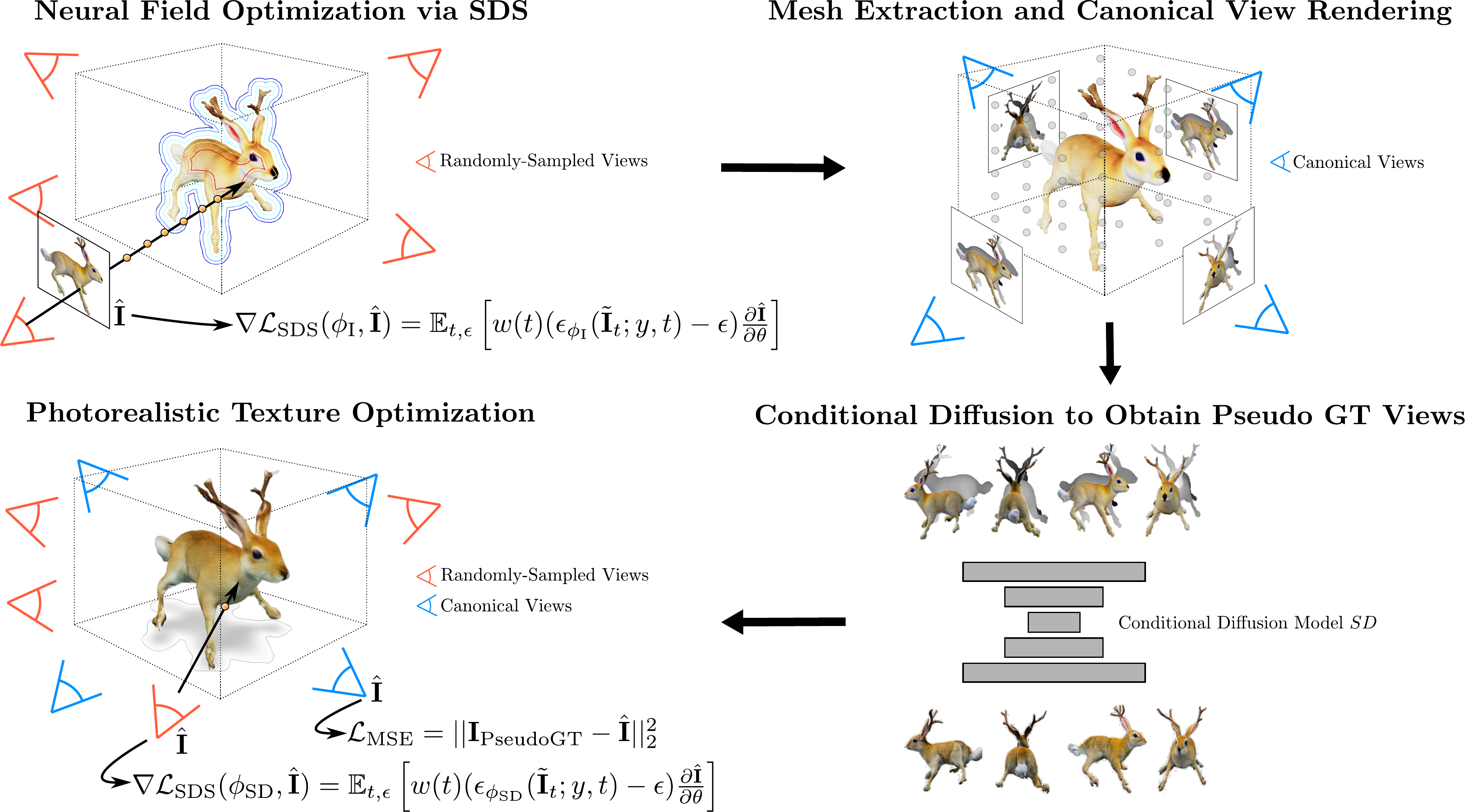}
    \caption{\textbf{Schematic Overview.} Given the input text prompt \emph{"an animal with the head of a rabbit, the body of a squirrel, the antlers of a deer, and legs of a pheasant"}, we train our initial distance field using Score Distillation Sampling (SDS) with view-dependent text prompting~\cite{poole2022dreamfusion} and an Imagen prior (top left) and extract the mesh with marching cubes (top right). 
    However, as the obtained appearance lacks details and the colors tend to be oversaturated, we  render the color and depth from four orthogonal views of our mesh (top right) and run them jointly through StableDiffusion to generate photorealistic and 3D consistent views of our mesh (bottom right). Eventually, we finetune the mesh texture on the obtained views together with a small SDS gradient to account for minor misalignments (bottom left).}
    \label{fig:method}
\end{figure*}

\section{Method}
Our goal is to develop a method that generates high-quality 3D mesh representations with photorealistic texture from text prompts.
In the following, we discuss the main components of our method. 
We first discuss our initial neural field-based geometry and appearance representation (\ref{subsec:initial_scene}) together with our first optimization stage where we train our model using a score-based distillation approach (\ref{subsec:initial-text-to-3d}).
Next, we describe how an initial mesh with texture can be extracted, and how we use this initial prediction to extract mesh-based RGB-D renderings.
This information is then used in our second optimization stage, where we combine the output of an image and depth-conditioned diffusion model together with a score-based distillation approach to obtain our final mesh with photorealistic texture as output (\ref{subsec:photorealistic-texture}). Fig~\ref{fig:method} presents an complete overview of our method. 

\subsection{Initial Scene Representation}\label{subsec:initial_scene}

\paragraph{Neural Radiance Fields}
A radiance field $f$ is a continuous mapping from a 3D location $\mathbf{x} \in \mathbb{R}^3$ and a ray viewing direction
 $\mathbf{d} \in \mathbb{S}^2$ 
to an RGB color $\mathbf{c} \in \left[ 0, 1 \right]^3$ and volume density $\sigma \in \reals^+$. In Neural Radiance Fields (NeRF)~\cite{Mildenhall2020ECCV}, this field $f_\theta$ is parameterized as a neural network with parameters $\theta$.
To render a pixel, a ray with direction $\mathbf{d} \in \reals^{2}$ is cast from the camera center and $M$ equidistant points $\mathbf{p}_i$ are sampled along the ray.
For a given camera pose $\mathbf{\xi} \in \mathbb{R}^{3 \times 4}$, the operator $\pi$ maps a pixel $\mathbf{u} \in \mathbb{R}^2$ to its color $\hat{\mathbf{I}} \in [0, 1]^3$ using classic volume rendering~\cite{Mildenhall2020ECCV}:
\begin{align}\label{eq:volume_rendering}
\pi: \left(\mathbf{\xi}, \mathbf{u}\right) \to \hat{\mathbf{I}}_u \,\,\,\text{,}\,\,\, \hat{\mathbf{I}}_u = \sum_{m=1}^{M} \alpha_m\mathbf{c}_m
\end{align}
where
\begin{align}
\quad \alpha_m &= T_m\left(1 - \exp(-\sigma_m\delta_m)\right)  \\
T_m &= \exp \left( -\sum_{m'=1}^{m} \sigma_{m'}\delta_{m'}\right) 
\end{align}
and $(\sigma_i, \mathbf{c}_i) = f_\theta(\mathbf{p}_i, \mathbf{d})$ are the evaluations along the ray and $\delta_i = \vert\vert \mathbf{p}_i - \mathbf{p}_j \vert\vert_2$ are the Euclidean distances between sampled points.

\paragraph{Signed Distance Fields} 
While NeRFs achieve impressive view synthesis results, the density-based representation is not well-suited for extracting a 3D geometry and obtaining a mesh~\cite{yariv2021volume}.
To overcome this limitation, we instead adopt an SDF-based representation:
\begin{align}
    f_\theta(\mathbf{p}_i, \mathbf{d}) = (s_i, \mathbf{c}_i) 
\end{align}
with $s_i \in \mathbb{R}$ being the signed distance from the surface at position $\mathbf{p}_i$. 
To enable training with volume rendering, we follow~\cite{yariv2021volume} and adopt the SDF to density transformation $t$:
\begin{equation}
    t_\sigma(s) = \alpha \Psi_\beta(-s),
\end{equation}
where
\begin{equation}
    \Psi_\beta(s) = 
    \begin{cases}
        \frac{1}{2} \exp\left(\frac{s}{\beta}\right) \quad\quad\quad \text{if}\, s \leq 0 \\ 
        1 - \frac{1}{2} \exp\left(-\frac{s}{\beta}\right) \quad \text{if}\, s > 0
    \end{cases}
\end{equation}
with $\alpha, \beta \in \mathbb{R}$ being learnable parameters. Using this transformation, our SDF-based neural field representation can be rendered to the image plane using the same volume rendering technique from \eqref{eq:volume_rendering}.

\subsection{Text-to-3D via Score-based Distillation}\label{subsec:initial-text-to-3d}
We generate our initial 3D model via training a neural distance field using a score distillation sampling approach. To this end, given a randomly sampled camera pose $\mathbf{\xi}$, we use our volume rendering operator $\pi$ from~\eqref{eq:volume_rendering} on all pixels $\mathbf{u}_i$ on the image plane to obtain the respective rendered image $\hat{\mathbf{I}}$\footnote{We drop the dependency on $\theta$, \ie  $\hat{\mathbf{I}} = \hat{\mathbf{I}}_\theta$, to avoid cluttered notation.}. We then sample random normal noise and time step $t$ and add it to the rendered image using two weighting factors $\alpha_t$ and $\sigma_t$ 
\begin{align}
\Tilde{\mathbf{I}}_t = \alpha_t \hat{\mathbf{I}} + \sigma_t \epsilon \quad\text{where}\quad
    \epsilon \sim N(0, I)
\end{align}
Following~\cite{poole2022dreamfusion}, $\sigma_t$ is chosen such that $\Tilde{\mathbf{I}}_t$ is close to the data density at the start of the diffusion process, \ie{,} $\sigma_0 \approx 0$ and converging to 1 for maximum diffusion steps, while $\alpha_t^2=1-\sigma_t^2$.
We feed $\Tilde{\mathbf{I}}$ to a diffusion model $\phi_{\text{I}}$ (Imagen~\cite{saharia2022photorealistic} in our experiments), which attempts to predict the noise $\epsilon$ with $\epsilon_{\phi_{\text{I}}}(\tilde{\mathbf{I}}, y, t)$,  given the noisy image $\Tilde{\mathbf{I}}$, diffusion step $t$, and text embedding $y$. From this prediction we can derive the gradient direction pushing the rendered images to a high probability density region for the provided text prompt with 
\begin{align}\label{eq:sds}
\begin{split}
    \nabla \mathcal{L}_{\mathrm{SDS}}(\phi_{\text{I}}, \hat{\mathbf{I}}) = \\
    \mathbb{E}_{t, \epsilon}\left[w(t)(\epsilon_{\phi_{\text{I}}}(\Tilde{\mathbf{I}}_t; y, t) - \epsilon) \frac{\partial \hat{\mathbf{I}}}{\partial \theta}\right],
\end{split}
\end{align}
where $w(t)$ denotes a weighting function and $y$ is the conditioning text embedding. This gradient is then utilized to optimize our signed distance field till convergence. Similar to \cite{lin2022magic3d} and \cite{poole2022dreamfusion}, we also employ classifier-free guidance~\cite{ho2022classifier} to control the strength of the text conditioning. Since this process involves rendering a MLP-based NeRF volume and running a pixel level diffusion model it can be carried out only at low resolution due to memory constraint. For this reason we compute $\mathcal{L}_{\mathrm{SDS}}$ on images rendered at 64$\times$64 (i.e., we use only the low resolution branch of Imagen). Note that the exact details for rendering, including shading and background modeling, match those from DreamFusion~\cite{poole2022dreamfusion}, which we omitted for the sake of clarity. We kindly refer to their paper for more information. However, unlike~\cite{poole2022dreamfusion}, we sample the whole elevation range for the camera, to avoid bleeding artifacts at the model bottom. 

Eventually, a mesh is extracted from the signed distance field as the surface at the zero-level set using Marching Cubes (MC)~\cite{lorensen1987marching}. Since floaters (\ie areas of near $0$ signed distance value away from the expected object surface) can occasionally remain within the volume, we additionally always select the largest mesh component closer to the center of the volume to create a mesh and to be used for the following steps.

\subsection{Photorealistic Texturing Using Multi-View Consistent Diffusion}\label{subsec:photorealistic-texture}
Upon having extracted the 3D mesh $\mathcal{M}$ from our trained distance field we already have a good geometry for the model. On the other hand, the texture still includes two main drawbacks: it misses high frequency details since the optimization in~\ref{subsec:initial-text-to-3d} is performed at low resolution only, and it shows over-saturated ('cartoonish'), colors as the result of using a large guidance weight~\cite{ho2022classifier}. %
To solve these two limitations we refine the initial texture using the standard pipeline of a Stable Diffusion model $SD$~\cite{rombach2022high} conditioned on color and depth. 
To this end, we take our obtained mesh, freeze its geometry, and use a differentiable-render $\mathcal{R}$ (NVdiffrast~\cite{laine2020modular}) to render color and depth from four canonical viewpoints $\mathcal{P}$ (i.e., \emph{front}, \emph{back}, and both \emph{sides}). Feeding the four views independently to a depth-conditioned diffusion model would be a straightforward way to obtain highly realistic images of the object, which could serve to guide the re-texturing. However, when processed independently, the resulting images exhibit several 3D inconsistencies, which would give the object a different identity depending on the viewpoint.

\begin{figure}
    \begin{subfigure}[b]{1.\linewidth}
    \centering
        \includegraphics[width=1.\linewidth,trim={0 1.cm 0 .5cm},clip]{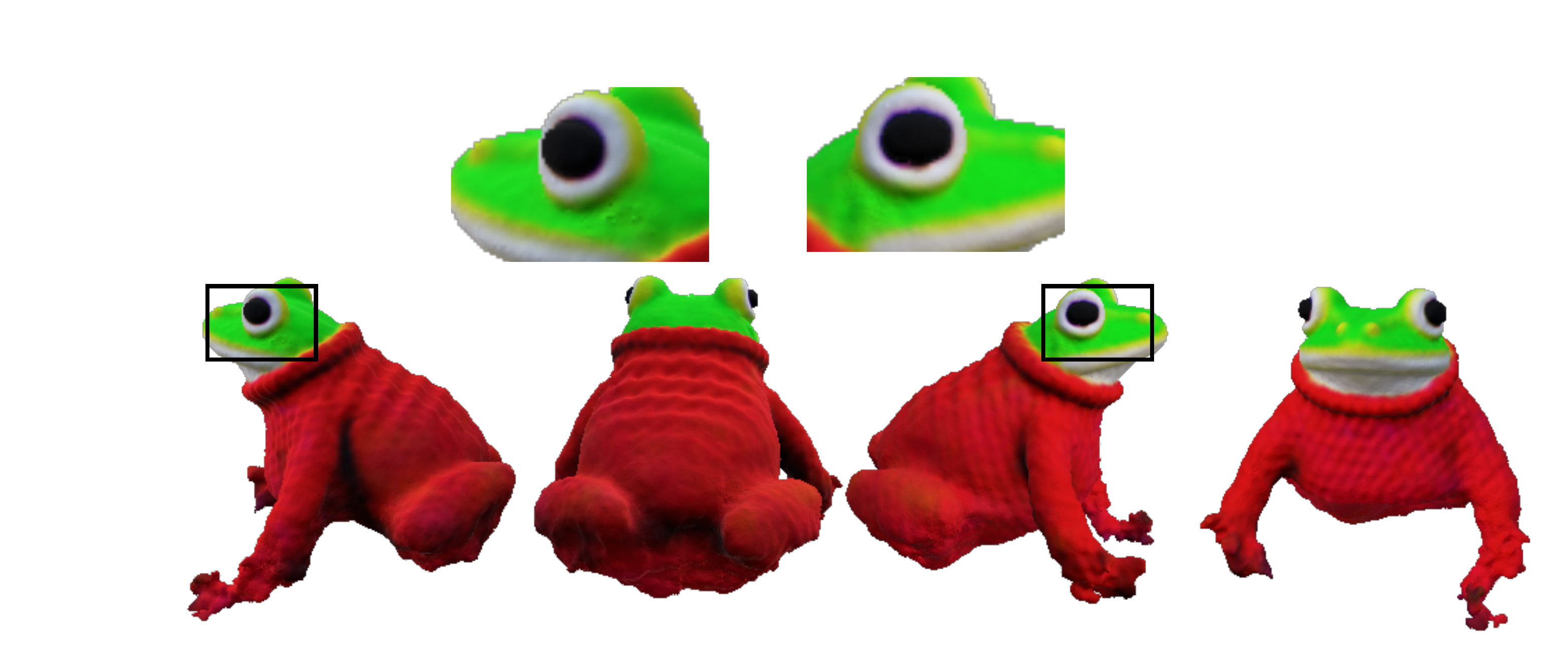}
        \subcaption{Input: Mesh rendering.}\label{subfig:tiling-a}
    \end{subfigure}
    \begin{subfigure}[b]{1.\linewidth}
    \centering
        \includegraphics[width=1.\linewidth,trim={0 1.cm 0 .5cm},clip]{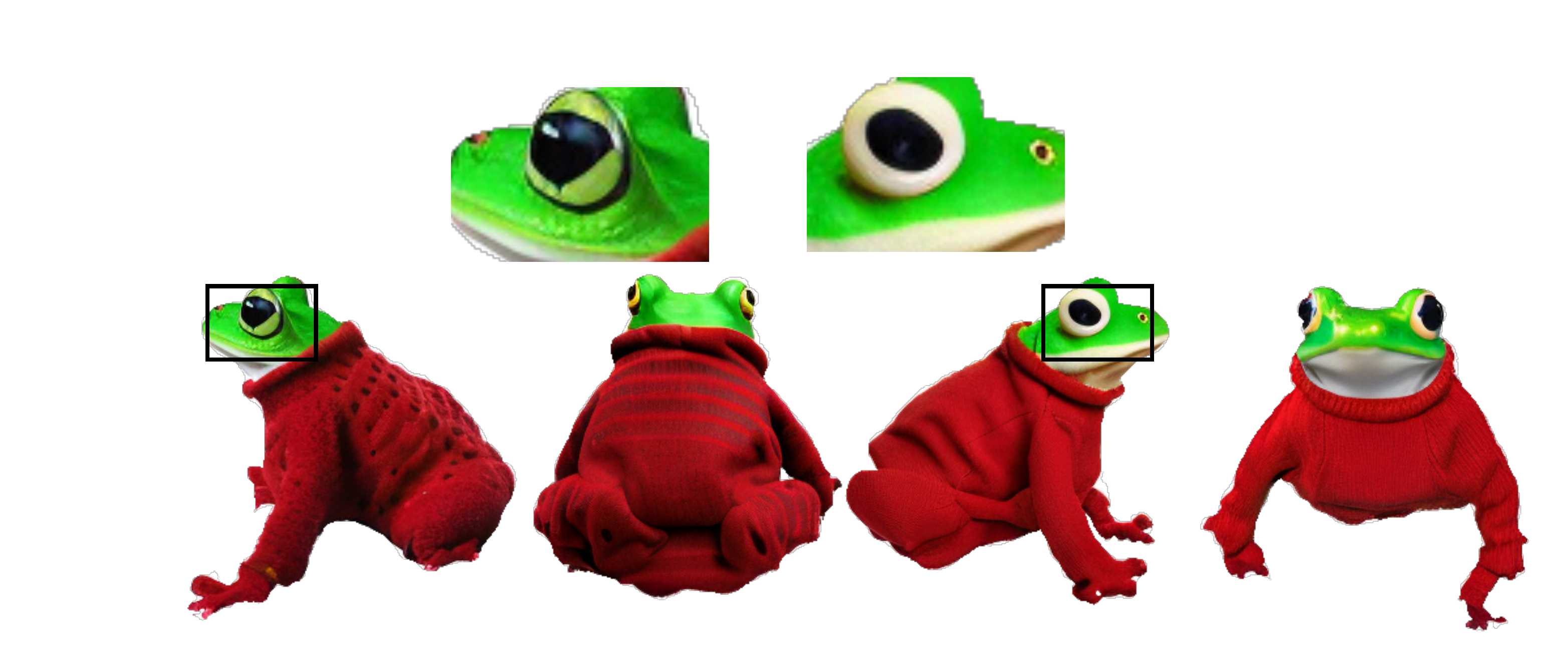}
        \subcaption{Output: Processed Independently.}\label{subfig:tiling-a}
    \end{subfigure}
    \begin{subfigure}[b]{1.\linewidth}
    \centering
        \includegraphics[width=1.\linewidth,trim={0 1.cm 0 .5cm},clip]{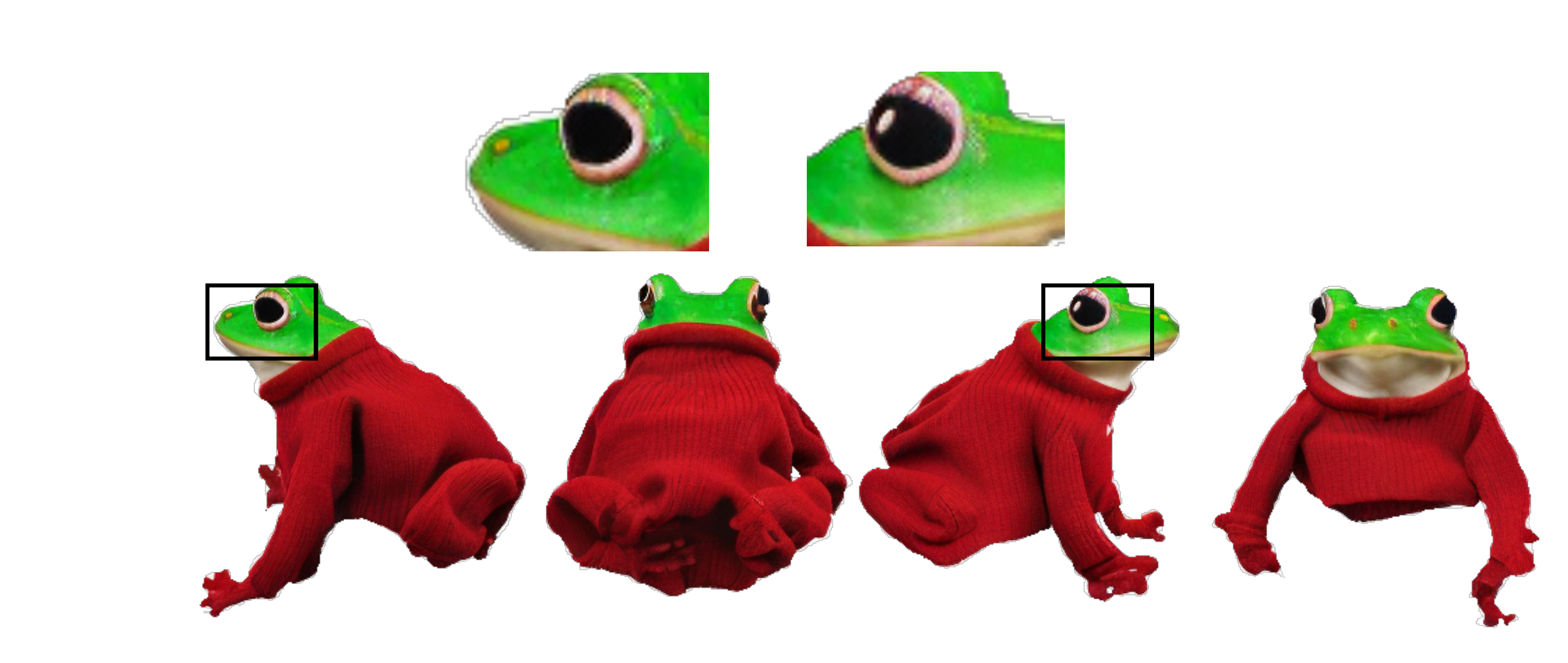}
        \subcaption{Output: Processed jointly (Ours).}\label{subfig:tiling-b}
    \end{subfigure}
    \caption{\textbf{Diffusion model conditioning strategies}. Input (a) and outputs of the depth conditioned diffusion model processing each input image independently (b) and conditioning  on all four input images jointly (c). We observe that the latter leads to multi-view consistent results, while processing each image independently can introduce inconsistencies (see zoom boxes).}
    \label{fig:tiled_views}
\end{figure}
To overcome this limitation, we propose to tile the four canonical RGB and depth predictions on a $2\times2$ grid to a single RGB image $\hat{\mathbf{I}}_\text{tiled}$ and depth map $\mathbf{D}_\text{tiled}$ and process them jointly in a single diffusion operation
\begin{equation}
    \mathbf{I}_\text{tiled} = SD(\hat{\mathbf{I}}_\text{tiled}, \mathbf{D}_\text{tiled}).
\end{equation} 
This enforces the diffusion model to generate consistent views during the diffusion of the tiled image (See Fig.~\ref{fig:tiled_views}). The individual pseudo GT views $\left\{\mathbf{I}_\text{PseudoGT, }i\right\}_{i=1}^{4}$ are extracted from image $\mathbf{I}_\text{tiled}$ and then serve as a pseudo ground truth %
that allows us to apply the new texture to the mesh geometry. 
The loss we optimize is
\begin{align}
   &\mathcal{L}_{\text{texture}}(\mathcal{R}, \mathcal{M}, P, i) =
         \vert\vert\mathbf{I}_{\mathrm{PseudoGT,}i} - \hat{\mathbf{I}}\vert\vert_2^2 \\
    &\text{with} \quad \hat{\mathbf{I}} =  \mathcal{R}(\mathcal{M}, P) \quad \text{for}\, P \in \mathcal{P} 
\end{align}

While tiling the views significantly improves 3D object consistency, the views can still exhibit minor misalignment at their intersection as well as on unobserved object parts. To ensure smooth transitions and a complete 3D mesh, we perform a second optimization stage, where we combine a photometric loss with a small SDS component using an image-to-image Stable Diffusion model $\phi_{\text{SD}}$. The new pseudo ground truth for this stage, $\{\mathbf{I}^\prime_{\mathrm{PseudoGT,}i}\}_{i}$, is obtained by rendering the converged texture from poses $\mathcal{P^\prime}$. %
In this stage, we then optimize
\begin{align}
  &\nabla\mathcal{L}_{\text{texture}}(\mathcal{R}, \mathcal{M}, P, i) = \nabla\mathcal{L}_{\mathrm{MSE}} + \lambda_{\mathrm{SDS}}\nabla  \mathcal{L}_{\mathrm{SDS}},\\
    &\mathrm{with} \quad \mathcal{L}_{\mathrm{MSE}} = \vert\vert\mathbf{I}^\prime_{\mathrm{PseudoGT},i} - \hat{\mathbf{I}}\vert\vert_2^2\\
    &\mathrm{and} \quad \hat{\mathbf{I}} =  \mathcal{R}(\mathcal{M}, P) \quad \text{for}\, P \in \mathcal{P^\prime} 
\end{align}
where $i$ describes the viewpoint for the camera poses $P \in P^\prime$ and and $\lambda_{\mathrm{SDS}}$ is the SDS weighting that controls its contribution to the texture optimization. For this step, we use a very small guidance weight of 7.5 compared to previous only SDS-supervised works, as it has been reported that increasing the guidance weight often results in saturated colors~\cite{ho2022classifier} and we only want to make small changes to the texture. Further, anchoring the optimization on $\mathbf{I}^\prime_{\mathrm{PseudoGT},i}$ enforces the resulting texture to not deviate too much from the original, encouraging only regions with high SDS gradients to change.

\section{Evaluation}

\subsection{Experimental Setup}

\paragraph{Metrics}
We follow previous work~\cite{jain2022zero, poole2022dreamfusion,mohammad2022clip} and evaluate our method using the CLIP~\cite{radford2021learning} R-Precision metric. This metric measures how well images rendered from the generated geometry correlate with the provided input text prompt; however, it fails to capture any aspect related to the 3D consistency and photorealistic appearance of the generated shapes. Therefore, we additionally report the $\mathrm{FID}_\mathrm{CLIP}$ score \cite{kynkaanniemi2022role}, which evaluates the FID in the feature space of the CLIP ViT-B-32 image encoder. As reference images, we use the ImageNet 2012 validation set. Similar to DreamFusion, we render 60 azimuthal angles at an elevation of 30 degrees to generate images which are used to evaluate both the R-Precision and $\mathrm{FID}_\mathrm{CLIP}$.

\paragraph{Text prompts}
We use 35 text prompts available on the public DreamFusion gallery\footnote{https://dreamfusion3d.github.io/gallery.html}, which mainly contains prompts describing individual objects making them suitable to be turned into meshes.

\subsection{Comparison with state-of-the-art}
\begin{table}[t!]
\begin{center}
\scalebox{0.89}{
\begin{tabular}{@{}lcccc@{}}
\toprule
            & \multicolumn{3}{c}{CLIP R-Precision $\uparrow$}   &  $\mathrm{FID}_\mathrm{CLIP}$ $\downarrow$\\  
Method      & B/32      & B/16      & L/14  & \\  \midrule
CLIP-Mesh   & 100       & 100       & 99.0       & 57.5\\
DreamFusion & 94.3      & 97.1      & 97.1       & 59.3\\  \midrule
Ours        & 91.4      & 91.4      & 94.3       & 57.4\\ \bottomrule
\end{tabular}
}
\end{center}
\caption{\label{tab:compre_DF_DF_CM}\textbf{Comparison with state of the art.} Comparing our method against the state of the art using R-Precision for different CLIP models and $\mathrm{FID}_\mathrm{CLIP}$. The metrics are computed on 35 prompts from the public DreamFusion gallery.}
\end{table}
\begin{figure*}
    \centering
    \includegraphics[width=\linewidth]{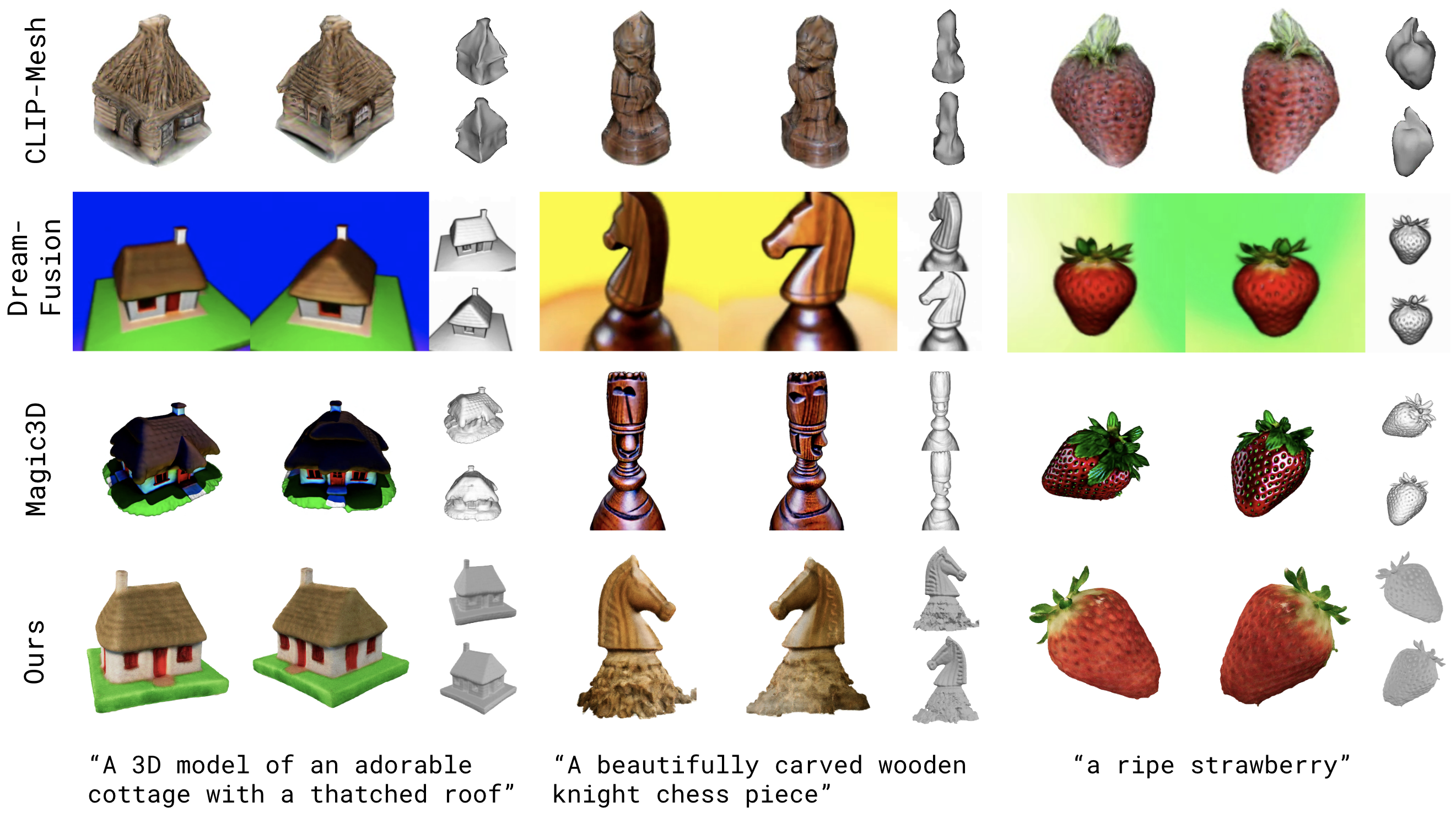}
    \caption{\textbf{Qualitative comparison} between our meshes, the meshes by CLIP-Mesh~\cite{mohammad2022clip} and Magic3d\cite{lin2022magic3d}, and the NeRF volumes by DreamFusion\cite{poole2022dreamfusion}. For each model we show both RGB renderings and 3D shape. Results for DreamFusion and Magic3D from \cite{lin2022magic3d}.}
    \label{fig:qualitative-comp}
\end{figure*}

We first compare our approach quantitatively against state-of-the-art methods in Table \ref{tab:compre_DF_DF_CM}. We compare to CLIP-Mesh as an alternative text-to-mesh generation model and to DreamFusion as a state-of-the-art text-to-NeRF method. For this comparison we re-run all the competitors on the same 35 prompts using the original code kindly provided by the authors and the default settings. Interestingly, although producing the qualitative worst results, CLIP-Mesh is able to report the best quantitative numbers for R-Precision. This is explained by the fact that CLIP-Mesh directly optimizes for the CLIP metric during its training. Our method performs on par with Dreamfusion, obtaining somewhat worse R-precision and better $\mathrm{FID}_\mathrm{CLIP}$ results. %

We also provide a qualitative comparison to the competitors in Figure \ref{fig:qualitative-comp}. For this comparison we also include Magic3D\cite{lin2022magic3d}, taking their results from their paper.
The qualitative comparison indicates that, while all methods allow for some degree of text-to-3D generation, our method achieves more photorealistic results. 
While all baselines tend to have a cartoonish and oversaturated appearance, we achieve more natural texture predictions thanks to our proposed texturing stage.
Note that a photorealistic appearance is crucial for many AR/VR applications where 3D content should fit in smoothly with the environment.

\subsection{Ablation Study}
\begin{table}[t!]
\begin{center}
\scalebox{0.95}{
    \begin{tabular}{@{}lcc@{}}
    \toprule
      &  R-Precision $\uparrow$ & $\mathrm{FID}_{\mathrm{CLIP}}$   $\downarrow$ \\  
    Method  & CLIP B/32 &. \\
    \midrule
    w/o texture finetuning     & 85.7  & 61.1\\
    w/o depth conditioning       & 91.4  & 57.7 \\
    w/o joint diffusion          & 91.4  & 55.9\\
    w/o multi-view loss          & 80.0  & 61.1 \\ \midrule
    Ours                         & 91.4  & 57.4\\ \bottomrule
\end{tabular}
}
\end{center}
\caption{\label{tab:ablations} \textbf{Ablation} of various components of our method using R-Precision and $\mathrm{FID}_\mathrm{CLIP}$ on 35 meshes. %
}
\end{table}
\begin{table}[t!]
    \centering
    \begin{tabular}{@{}lc@{}}
         Criteria & Preference $(\%)\uparrow$ \\ \midrule
         More Natural Colors & 61.2 \\
         More Detailed Texture & 63.3 \\
         Overall Visually Preferred & 57.9 \\ \bottomrule
    \end{tabular}
    \caption{\textbf{User Study.} Results of our user study, conducted with 30 participants. Each participant was shown two meshes, before and after our re-texturing using depth-condition diffusion, for a total of 15 prompts from the DreamFusion~\cite{poole2022dreamfusion} gallery and had to choose which mesh they preferred based on different criteria.}
    \label{tab:user_study}
\end{table}

In Table \ref{tab:ablations} we ablate various components of our method. Each ablation is evaluated on 35 prompts and, as before, we report the CLIP R-Precision and $\mathrm{FID}_\mathrm{CLIP}$ score. In our default setting, we use a depth-conditioned Stable Diffusion model and four views of the mesh seen from the front, back and sides (see~Figure \ref{fig:method}). The views are then tiled in a grid and processed as one image by Stable Diffusion.

\begin{figure*}[t]
    \centering
    \includegraphics[width=.9\linewidth]{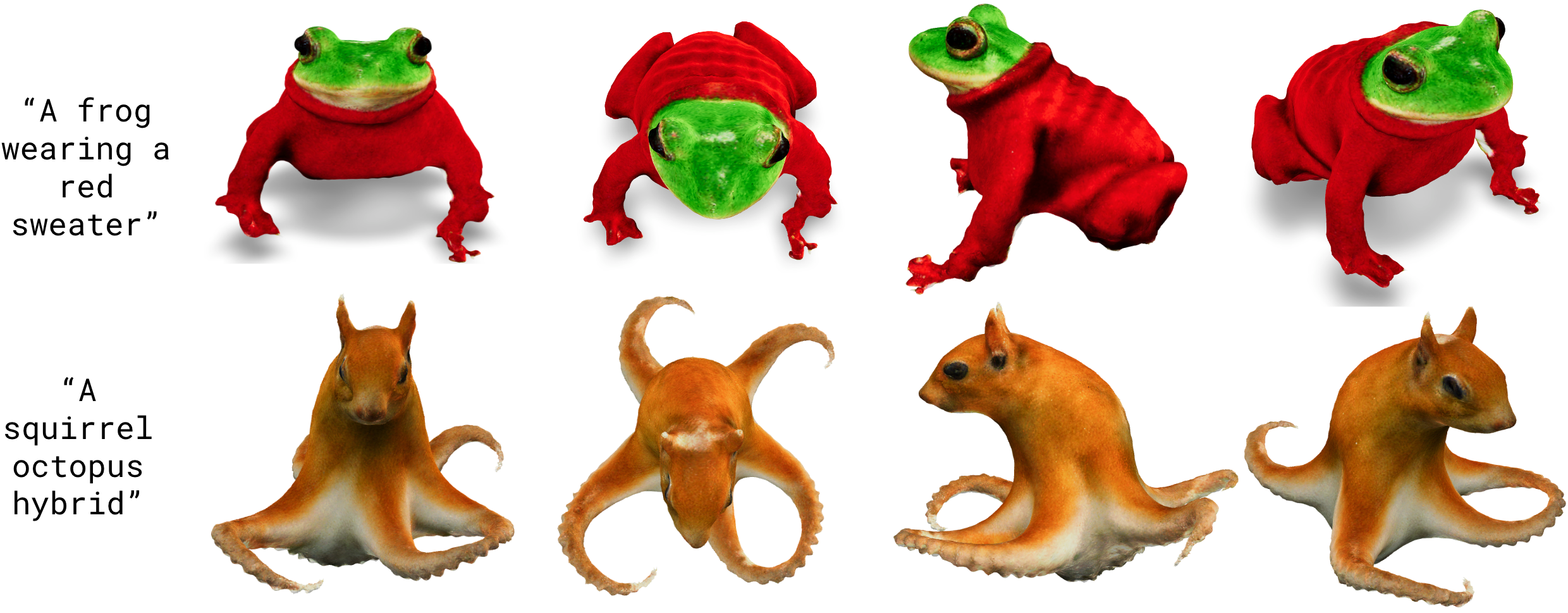}
    \caption{\textbf{3D Consistency.} Our depth-conditioned joint diffusion process ensures that obtained meshes are geometrically accurate and consistent in 3D space.}
    \label{fig:3d-cons}
\end{figure*}

\paragraph{Quantitative Results} First, the results in Table~\ref{tab:ablations} indicate that the texture finetuning stage is crucial to obtain realistic-looking meshes, as all refined options obtain better R-Precision and $\mathrm{FID}_\mathrm{CLIP}$ scores. Further, when removing depth conditioning or joint diffusion, we obtain overall similar results, with separate diffusion obtaining slightly better results for $\mathrm{FID}_\mathrm{CLIP}$. We attribute this to the fact that the employed metrics, including $\mathrm{FID}_\mathrm{CLIP}$, are not very suitable at evaluating the 3D consistency of the generated texture. We kindly refer to the supplement, where we provide several examples demonstrating that removing joint diffusion leads to inconsistencies in 3D space. Finally, the multi-view component is essential for obtaining realistic results, as the SDS-only driven optimization performs worst overall. %

\paragraph{User Study} While the reported metrics can provide an indication of the quality of the results, it is important to note that they do not directly measure the perceived quality of the generated models.
To further quantify the importance of our texture finetuning stage, we perform a user study comparing the results before and after the photorealistic texturing stage in Table \ref{tab:user_study}.
We observe that humans prefer the results after the texturing stage, in particular with respect to texture details and color.%

\subsection{Mesh Quality}
\begin{figure}[t!]
    \centering
    \includegraphics[width=0.95\linewidth]{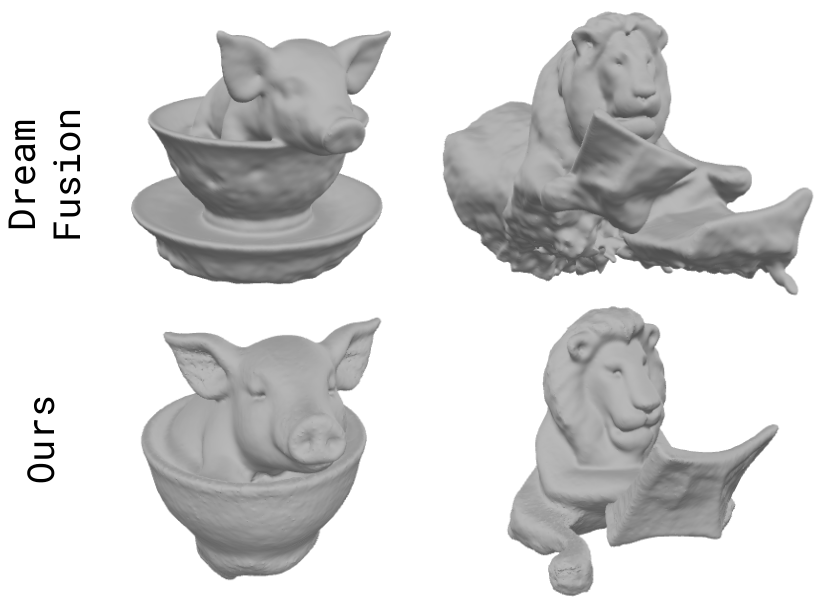}
    \caption{\textbf{Comparing the 3D mesh geometry} from the radiance field of DreamFusion and our SDF-based approach.}
    \label{fig:mesh_qual_results}
\end{figure}

\paragraph{Extracting Meshes}
In Figure \ref{fig:mesh_qual_results}, we compare extracted meshes from Dreamfusion~\cite{poole2022dreamfusion} and our method.
We observe that our SDF-based approach leads to smoother mesh predictions.
Obtaining a high-quality mesh as output representation is crucial for many applications, and we provide an example for an AR application in Figure \ref{fig:teaser}. To generate the meshes for DreamFusion we use marching cubes over the volumes obtained for the evaluations in Table~\ref{tab:compre_DF_DF_CM}, i.e., applying the default settings provided by the authors. Besides the use of a SDF volume, a key difference between our method and Dreamfusion is that we sample the full elevation range of camera poses while DreamFusion uses a limited one. This allows us to obtain nice complete meshes without spurious surfaces (e.g., the artefacts on the bottom of the DreamFusion lion).

\paragraph{3D Consistency}

In Figure \ref{fig:3d-cons}, we show multiple views for the same object to provide a qualitative evaluation of the 3D consistency of the optimized meshes with photorealistic texture.
We find that the final outputs of our method are indeed 3D consistent and that the texture appears realistic from arbitrary viewpoints. 
\section{Conclusion}

We present TextMesh, a novel approach for 3D mesh generation from text prompts. In the core, we propose to represent the geometry as a distance field which is optimized using Score Distillation Sampling (SDS). After optimization of the distance field, we then extract the mesh and refine its original texture to achieve a more detailed and natural appearance. In contrast to similar methods, we supervise the texture refinement primarily with a photometric loss on enhanced 2D mesh renderings generated by a depth condition image-to-image diffusion model, and only rely on SDS to smooth out transitions within our multi-view supervision. This leads to more photorealistic textures, preferred by a larger portion of our survey participants compared to the initial, unrefined texture.%

\FloatBarrier

{\small
\bibliographystyle{ieee_fullname}
\bibliography{bib_long,bib,egbib}
}

\end{document}